\documentclass[10pt,twocolumn,letterpaper]{article}

\usepackage{iccv}
\usepackage{times}
\usepackage{epsfig}
\usepackage{graphicx}
\usepackage{amsmath}
\usepackage{amssymb}

\usepackage{latexsym}
\usepackage{url}
\usepackage{epsfig}

\usepackage{algorithm}
\usepackage[noend]{algpseudocode}

\makeatletter
\def\BState{\State\hskip-\ALG@thistlm}
\makeatother

\usepackage[breaklinks=true,bookmarks=false]{hyperref}

\iccvfinalcopy 

\def\iccvPaperID{****} 

\ificcvfinal\fi

\begin{document}


\title{Granular Multimodal Attention Networks for Visual Dialog}

\author{ {Badri N. Patro} \quad {Shivansh Patel}  \quad {Vinay P. Namboodiri} \\
Indian Institute of Technology, Kanpur \\
{\tt\small \{ badri,shivp,vinaypn \}@iitk.ac.in}
}

\maketitle
\ificcvfinal\thispagestyle{empty}\fi


\begin{abstract}
Vision and language tasks have benefited from attention. There have been a number of different attention models proposed. However, the scale at which attention needs to be applied has not been well examined. Particularly, in this work we propose a new method Granular Multi-modal Attention, where we aim to particularly address the question of the right granularity at which one needs to attend while solving Visual Dialog task. The proposed method shows improvement in both image and text attention networks. We then propose a granular Multi-modal Attention network that jointly attends on the image and text granules and shows the best performance. With this work, we observe that obtaining granular attention and doing exhaustive Multi-modal Attention appears to be the best way to attend while solving visual dialog.
\end{abstract}
\vspace{-1.8em}
\section{Introduction}
\vspace{-0.5em}
In `Visual Dialog' \cite{Das_CVPR2017} problem, an AI agent has access to an image. The aim is that the bot should be able to answer a question given the image and the context of the previous conversation. We can gain insights for a method by observing the regions of the image the method most focuses on while answering a question. It has been observed in a recent work that humans also attend to specific regions of an image while answering questions \cite{Das_EMNLP2016}. We therefore expect strong correlation between focusing on the ``right’’ regions while answering questions and obtaining better semantic understanding to solve the problem. This correlation exists as far as humans are concerned~\cite{Das_EMNLP2016}. We therefore aim in this paper to obtain image based attention regions that correlate better with human attention. It is known that using attention for solving various problems that relate to vision and language is a good approach. However, in an interesting evaluation carried out for the visual question answering task \cite{Das_EMNLP2016} it was observed that the attention networks focus on regions different from that used by humans for answering questions. We hypothesize that this could be due to the fact that they do not focus on the right granularity and correct context when obtaining attention for image and textual regions.

\begin{figure}[ht]
\centering
\includegraphics[width=0.5\textwidth]{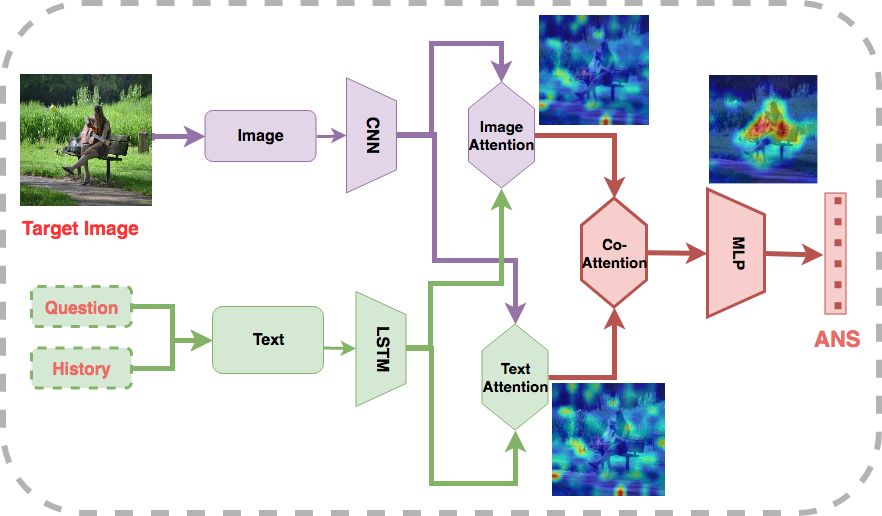}
\caption{ We first obtain image and text attention. Then we used Multi-modal Attention network to obtain final attention map. Finally, we classify the answer based on the attended feature. We provide the attention map that indicates the actual improvement in attention.}
\label{fig:moti}
\end{figure}
In this work, we aim to address this problem. We particularly aim to obtain  `granular regions' for images using object proposals as was used earlier by Anderson {\it et al.} \cite{anderson_CVPR2018bottom} and word based attention using the appropriate context. We observe that by using all the text to attend to each image proposal using granular image attention network and using the whole image to attend to each word using granular word attention networks, one can obtain the appropriate attention network regions in image and text respectively. Further, these when used in conjunction result in further improvement through granular multi-modal attention networks. In figure~\ref{fig:moti} we illustrate the main idea of the network. As can be observed, the multi-modal attention network obtains attention regions from both image and text and when combined provides improved attention regions. 

As part of the work, we also carry out a thorough evaluation of all the main attention methods that have been proposed in literature for various vision and language tasks. This evaluation also provides the ground for properly analyzing the various attention methods. Additionally, we consider the correlation between the visual explanation and attention regions. In literature, there have been visualization efforts such as Grad-cam \cite{selvaraju2017grad} that aim to provide explanations for the decisions by neural networks by visualizing the gradient information. We show that the proposed attention network regions correlates well with these regions obtained using Grad-cam. To summarize through this paper we provide the following main contributions:

\begin{itemize}
    \item We propose a granular image-attention (GIA) and granular text-attention (GTA)  based approach to obtain improved attention regions for solving visual dialog.
    \item We evaluate three variants of the proposed granular attention networks - one where we only obtain image attention, the other where we obtain text based image attention and the final proposed method where we combine these attentions using multi-modal attention method. 
    \item We obtain an improved overall accuracy by ~6\% NDGC score as compared to other baseline approaches using the proposed attention model for the visual dialog task.
  \item We provide a thorough empirical analysis for the method and also provide visualizations of attention mask for granular multi-modal attention network(GMA) and measure rank correlation among various attention masks with Grad-CAM regions to ensure that the attention regions correlate with the regions used by network in solving visual dialog.
\end{itemize}


\section{Related Work}
\label{sec:lit_surv}
A conversation about an image is known as Visual Dialog. This is one of the recent challenges in the field of vision and language. The task of Visual dialog involves image captioning ~\cite{Socher_TACL2014, Vinyals_CVPR2015, Karpathy_CVPR2015, Fang_CVPR2015, Johnson_CVPR2016, Yan_ECCV2016} that is description about image, visual question answering~\cite{Malinowski_NIPS2014, VQA, Ren_NIPS2015, Noh_CVPR2016, Xu_ECCV2016, Shih_CVPR2016, Lu_NIPS2016} that is responding natural language question about an image, visual question generation ~\cite{Mostafazadeh_ACL2016,Patro_EMNLP2018MDN} that generating natural language question about an image and generating similar types of question of given question\cite{Patro_COLING2018learning}. The multimodal attention mechanism is one of the core mechanisms in the interaction system. In VQA,\cite{Zhu_CVPR2016} proposed attention network in stacked fashion,\cite{Fukui_arXiv2016} combine both modality in frequency space and \cite{Patro_CVPR2018dvqa} use exemplar way to combine and get multimodal attention map. \cite{hajivc2017visual} proposed a cross-modal attention network by looking at word level and object level. \cite{Patro_ICCV2019} proposed a nice algorithm for minimizing uncertainty and get a robust attention map in a multimodal system.  Visual dialog requires the agents to have a meaningful conversation about the visual content of an image. This was introduced by \cite{Das_CVPR2017}. They proposed three approaches, i.e, late fusion, by concatenating all the history around, attention-based hierarchical LSTM for handing variable-length history and memory-based method which resulted in the highest accuracy. 
 
Das \textit{et al.}\cite{Das_ICCV2017} proposed deep reinforcement learning-based end-to-end trained model for Visual Dialog. Strub \textit{et al.}\cite{strub2017end} proposed an end-to-end RL optimization and it's applications to multimodal tasks. Chattopadhyay \textit{et al.}\cite{visdial_eval} designed an interactive AI image guessing game on visual dialog. Lu \textit{et al.} \cite{lu2017best} and Wu \textit{et al.}\cite{wu_CVPR2018} proposed generator and discriminator based architecture. Recent work on the visual dialog as proposed by Jain \textit{et al.}\cite{jain2018two} is based on discriminative question generation and answering. \cite{Patro2019ProbabilisticFF} proposed a probabilistic method to generate diverse answers and also minimize uncertainty in the answer generation. Various methods have been proposed to handle variable-length history rounds. Das \textit{et al.}\cite{Das_CVPR2017} has also proposed the `Late Fusion' (LF) method, where question word tokens are concatenated with answer tokens then obtained its embedding using LSTM network. In order to handle variable-length history, they proposed hierarchical LSTM. Also, they proposed a memory network model to perform the best results in terms of accuracy. In contrast to the earlier architectures, we address the question of obtaining correct attention regions for solving the task and provide comparisons with the related attention methods.


	\begin{figure}[htb]
	\centering
	\includegraphics[width=0.4\textwidth]{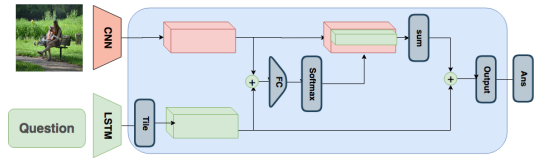}
	\caption{Illustration of SAN and its attention mask. We pass an image through a convolutional neural network to get image features and apply tile function on text to obtain text features. Then we obtain softmax mask using attention network. We apply the attention mask on image to get the final attention.}
	\label{fig:background-1}
\end{figure}
\section{Background: Stacked Attention Network (SAN) }	

		Stacked Attention Network (SAN)~\cite{Yang_CVPR2016} is a question guided attention scheme which learns the attention probability vector of the visual information based on the input question vector. We use this as our reference network. Attention is weighted average of question features and image features. The output of weighted average features is fed into tanh layer followed by linear layer to compute the attention probability vector. Softmax is applied over linear layer to obtain attention map, which indicates the probability of contribution of each spatial feature. Finally, this attention map is multiplied by image feature and resultant is added to question feature to predict answer. SAN uses stack of this attention layer as a iterative step to narrow down the selection portion of visual information. Mathematical expression for SAN is as follows:
	\[f^q_{j}=Att(f_q^{j-1},f_{i}) + f_q^{j-1}\]
	where $f_q^{j}$ is the question vector in $j^{th}$ iteration and $f_{i}$ is the image feature matrix. This process repeats $J$ times to obtain correct answer.
	\[{Answer}=\text{softmax}{(W*f^q_{j} +b)}\]

\subsection{Multimodal Compact Bilinear Pooling(MCB)}
	\begin{figure}[htb]
	\centering
	\includegraphics[width=0.45\textwidth]{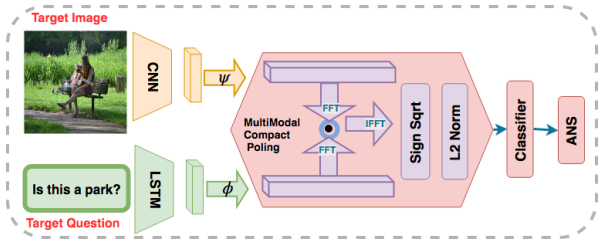}
	\caption{ Illustration of Multimodal Compact Bilinear Pooling(MCB) and its attention mask. We obtain image features using Convolution Neural Network and question features using LSTM. Finally, we combine these two using Fast Fourier Transform and obtain final attention vector by applying Inverse Fast Fourier Transform to final attention. Then we normalize the attention vector and obtain the final answer}
	\label{fig:background-2}
\end{figure}

	MCB is another approaches to combine image modality with language modality. Here, we obtain image features using Convolution Neural Network and text features using Recurrent Neural Network. Finally, we combine these two using Fast Fourier Transform and obtain final attention vector by applying Inverse Fast Fourier Transform to final attention. Then we normalize the attention vector and obtain the final answer. In compact bi-linear pooling, we transform image features and question features to common space using compact bi-linear pooling. From this, we obtain correlation feature between image features and question features. Then we obtain attention mask by weighted average of question features and image features. The output of weighted average features is fed through tanh layer followed by linear layer which makes their sizes equal. Soft-max is applied over linear layer to get attention map which indicate the probability of contribution of each image feature. Finally, this attention map is multiplied by ``conv5'' features and resultant is added to question features to predict answer.

\begin{equation}
    \begin{split}
        & f^{iq}=\tanh((W_{c}f^i_{n,m}+b_{c}) \odot W_{q} f^q)\\
        & f_{ls}^{iq}=L_{2}(\text{signedsqrt}(f^{iq})\\
        &f_{a}^{iq}=W_{a2}\sigma(W_{a1}f_{ls}^{iq} +b_{a1})\\
        & \alpha_{n,m} = \text{softmax}(f_{a}^{iq}) \\
        & f_{att} = \sum_{n=1}^{N}\sum_{m=1}^{N}{\alpha_{n,m}*f^i_{n,m}} \\
        & f_{out} = f_{att}  \odot f^q
    \end{split}
\end{equation}
	
	\begin{figure*}[htb]
	\centering
	\includegraphics[width=1.05\textwidth]{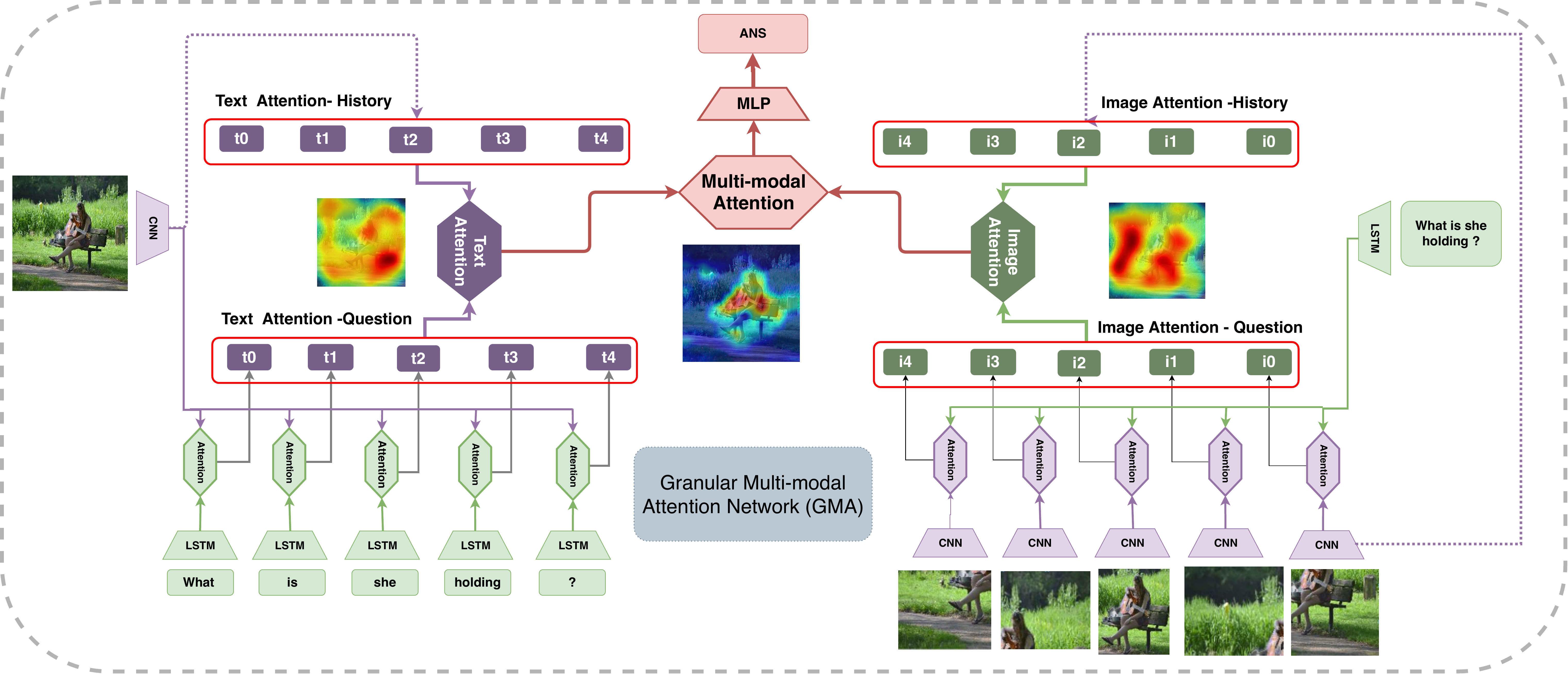}
	\caption{ Granular Multi-modal Attention Attention Network. Firstly, we use a CNN to obtain image features $g_i$, then we use LSTM to obtain text features. We obtain image attention features by attending question features with each object present in the image. Similarly, we obtain text attention feature by attending image features with every word in question. Afterwards, we use a Multi-modal Attention network to obtain final attention. Final answer can be obtained using the result of Multi-modal Attention.}
	\label{fig:main_figure}
\end{figure*}
\section{Method}
The main focus in our approach for solving visual dialog task is to use multi-modal attention based method to  combine granular image and textual attention map to improve attention and visual explanation. The key differences in our architecture as compared to an existing visual dialog architecture is the use of granular Attention mask.  The other aspects of dialog are retained as is. In particular, we adopt a classification based approach for solving visual dialog where an image embedding is combined with the question  and history embedding to solve for the answer.  This is illustrated in figure~\ref{fig:main_figure}.
\subsection{Overview}
Our method consists of four parts as illustrated in figure~\ref{fig:main_figure}:

\begin{enumerate}
	\item We obtain granular feature for image and Question embeddings using standard pretrained VQA (MCB) Model\cite{fukui2016multimodal}.
	\item  We obtain granular image attention feature for question and history based on query image. Similarly, we obtain granular text attention feature for the text based on the query question and previous history.
	\item Then, we combine these attention mechanisms  using a  Multi-modal Attention mechanism. We also evaluate Multimodal Compact Bilinear Pooling mechanism and concatenation to combine these two attentions but we observed that Multi-modal Attention performs better. Finally, we obtain context encoding vector.
	\item This context embedding compares with all the candidate answer options and obtain a 100 dim vector by reasoning with each answer option and context embedding vector. 
\end{enumerate}

 
\subsection{Granular Multimodal Attention Module }
Granularity in image and question is required to answer each round of question in visual dialog. 
It is built on two main networks - Granular Image Attention Network and Granular Text Attention Network. In image attention network, we obtain attention features by attending using text features for each object present in image.  In text attention, we obtain attention features by attending image features for each question (history token). The joint attention model (GMA) combines image and text attention features to obtain final answer encoding feature.

\subsubsection{Granular Image Attention (GIA) module}
\label{ssec:imgmaskgeneration}
In Granular Image Attention Network, we obtain relevant regions in an image for answering particular question in visual dialog. \cite{petsiuk2018rise} generates importance regions indicating how salient each pixel is for the model’s prediction. Following this explanation approach, we mask different image regions in random combinations to compute the importance of each region to produce the ground truth answer. Saliency map for a given output class is computed as a weighted sum of random masks, where weights are the probability scores of that ground-truth class for the corresponding mask. We generate saliency map for the ground-truth class.  We use the image features from vgg-16,CONV-5 given by $X \in \mathcal{R}^{C\times N}$ where $N=14\times 14$ and $C$=512. To produce an importance map, we randomly sample a set of binary masks ${M_1,...,M_N}$ according to distribution $D$ and probe the model by running it on masked image regions $X \odot M_i$, $i = 1,...,N$ and $\odot$ denotes element-wise multiplication. Then, we take the weighted average of the masks where the weights are the confidence scores $f(X \odot M)$ of the ground-truth class and normalise it by the expectation of $M$. 
\begin{equation}
    \begin{split}
    S_{X,f} = \frac{1}{E[M].N}\sum_{i=1}^{N}f(X \odot M_i)M_i 
    \end{split}
\end{equation}
Where $S_{X,f}$ denotes the saliency map for the ground-truth class. The map generated denotes the importance of each image region for predicting the ground truth class and therefore, we use this map obtained as a basis for improving the image-attention network of the baseline model.  We treat each round of the visual dialog as a visual question answer (VQA) \cite{wu_PAMI2017image} with external knowledge as a history as a context input. So We obtain image saliency map for each round of the visual dialog. We term this as granular image feature. The attention applied on this image  feature is know as granular image attention.  Then we apply attention mask to get attention probability of each of the spatial region. The entire procedure is as follows:
\begin{equation}
    \begin{split}
        & f^{iq}=\tanh((W_{c}f^i_{n,m}+b_{c}) \odot W_{q} f^q)\\
        & f_{ls}^{iq}=L_{2}(\text{signed\_sqrt}(f^{iq}))\\
        &f_{a}^{iq}=W_{a2}\sigma(W_{a1}f_{ls}^{iq} +b_{a1})\\
        & \alpha_{n,m}^{q} = \text{softmax}(f_{a}^{iq}) \\
        & {f^{'}_{q}} = \sum_{n=1}^{N}\sum_{m=1}^{N}{\alpha_{n,m}^{q}*f^i_{n,m}} \\
    \end{split}
\end{equation}
where $f_{n,m}^i$ is a vector representing $n \times m$ bounding box. $f_q$ is a vector representing the question, $(W_{*}, b_{*})$, are weight matrices and bias terms. $\alpha^{q}_{n,m}$ is the attention probability mask. $f^{'}_{q}$ is the image attended feature for question sequence. Similarly, we obtain image attention feature vector $f^{'}_{h}$ for history sequence. Finally we combine image attended  feature for question $f^{'}_{q}$ with  image attended  feature for history $f^{'}_{h}$ to obtain final image attended feature $A_i$ which is shown in algorithm ~\ref{GMA}

\subsubsection{Granular Text Attention (GTA) module}
\label{ssec: questionmaskgeneration}
Similar to \ref{ssec:imgmaskgeneration}, We find out the question words that the model needs to attend more in order to predict the answer properly. Suppose we have a question $Q$, which has $n$ words $(q_1, q_2, q_3,......,q_n)$. We mask two words at a time and probe the model to run with the masked question $Q'$ . For eq , $Q'$ can be $(q_1, q_2, 0 ,0,.........,q_n)$ . There are total $^{n}C_{2}$ possible combinations for $Q'$. 
Let the ground truth class probability predicted by the  model when probed with masked Question be $G'$. We find the masked question $Q'$ with which we get the maximum value of $G'$ and further, retrieve the attention map $\hat Att_q$ generated by the model when probed with $Q'$. Then we multiply the attention probability of each word with each question token which is as follows: 
\begin{equation}
    \begin{split}
        C^k &= \tanh{(W_{i}g_i+b_{i} +W_{q}f^k_{q}+b_{q})} \\
        p^{k} &= softmax(W_{c}C^{k} +b_c) \\
        f^{'}_{q} &= \sum_{k=1}^{T}{p^{k}f^k_{q}}
    \end{split}
\end{equation}
where $\hat{g}_i$ is a vector representation which brings $g_i$ image feature to a common feature space. ${\hat{f}_{q}}^k$ is a vector representing word $k$ which brings $f_q^k$ feature to a common feature space, where $k \in 1,..., K$, $K$ is the length of the sequence and $p^{k}$ is the attention probability mask. Here $(W_{*}, b_{*})$ are weight matrices and bias terms. $f^{'}_{q}$ is the attended question feature for question sequence. Similarly we obtain attention feature vector $f^{'}_{h}$ for history sequence. Finally we combine attended question feature with attended history feature to obtain final text attended feature $A_t$ which is shown in algorithm ~\ref{GMA}.

\subsubsection{Granular Multi-modal Attention (GMA)}
The challenging task of this module is to combine image attention (GIA) and text attention (GTA) modality. We adapt most efficient method, Multimodal Compact Bilinear Pooling (MCB) ~\cite{fukui2016multimodal}, to combine image modality with language modality. This method uses Fast Fourier Transform to convert image  and text space into Fourier space, where it combine both modality using compact bi-linear pooling method, then using  Inverse Fast Fourier Transform to bring back to final attention space. 
From this, we obtain correlation feature between image features and question features. Then we obtain attention mask by weighted average of question features and image features. The output of weighted average features is fed through tanh layer followed by linear layer which makes their sizes equal. Softmax is applied over linear layer to get attention map which indicate the probability of contribution of each image feature. Finally, this attention map is multiplied by image features to predict final answer which is as follows: 
\begin{equation}
    \begin{split}
        C_A &={MCB}_{16,000}(A_{i},A_{t})\\
        \gamma &= softmax(W_{c}C_A + b_{c}) \\
        A &= \sum{\gamma g_{i}}
    \end{split}
\end{equation}
where $A_i, A_{t}$ are the attended vector representing for GAIN  and GTAN. $(W_{*}, b_{*})$ are weight matrices and bias terms. $\gamma$ is the attention probability mask and $A$ is the final attention vector. In MCB method, we first project the lower dimensional inputs $A_i$ and $A_t$ to a 16,000 dimensional space and combines them and later projects it back into a lower dimensional space to get a joint feature representation $C_A$. Finally we minimize the cross entropy loss over all training examples. The cross entropy loss between the predicted and ground truth answer is given by:
\begin{equation}
 L_{cross}( s, y) = -\frac{1}{C}\sum_{j=1}^{C} y_{j} log(c_{j}|s) 
 \end{equation}


\begin{algorithm}
	  \scriptsize
	\caption{Granular Multi-modal Attention}\label{GMA}
	\begin{algorithmic}[1]
		
		\State I: Given input image
		 \State Q: 10 rounds of Question
		  \State H: 10 rounds of Question and Answer pair
		 
		\BState \emph{\textbf{GMA Mechanism}}:
		\While{loop} 
		\State Compute Image Embedding $g^k_i=G(I,W_{c})$
		\State Compute Question Encoding  $f^k_q=F(Q,W_{q})$
		\State Compute History Encoding  $f^k_h=F(H,W_{h})$
		\While{k=1:K} (For GTAN)
		   \State $\alpha^k_q$ = ATTENTION $(g_i, f^k_q)$
		   \State $\alpha^k_h$ = ATTENTION $(g_i, f^k_h)$
		   \State  ${f^{'}}^k_q =\sum{\alpha^{k}_{q}*f^k_q}$
		   \State  ${f^{'}}^k_h =\sum{\alpha^{k}_{h}*f^k_h}$
		   \State $\alpha^k$ = ATTENTION $({f^{'}}^k_h, {f^{'}}^k_q)$
		   \State  ${A}^k_t =\sum{\alpha^{k}*f^k_q}$
		   \EndWhile
		\While{l=1:L}  (For GIAN)
		   \State $\beta^l_q$ = ATTENTION $(g^l_i, f_q)$
		   \State $\beta^l_h$ = ATTENTION $(g^l_i, f_h)$
		    \State  ${f^{'}}^l_q =\sum{\beta^{k}_{q}*g^l_i}$
		     \State  ${f^{'}}^l_h =\sum{\beta^{k}_{h}*g^l_i}$
		     \State $\beta^l$ = ATTENTION $({f^{'}}^l_h, {f^{'}}^l_q)$
		   \State  ${A}^k_i =\sum{\beta^{k}*g^k_i}$
		   \EndWhile
		   
		 \State $\gamma_{gma}$ = MCB\_ATT$({A}_i, {A}_t)$
		 \State  $ A=\sum{\mu_{\small{gma}}*g_i}$
		\EndWhile
		\State {-------------------------------------------------------}		
		
		 \Procedure{:attention}{$g_{i}$, $f_{i}$}
		\State Image feature: $g_{i} \in R^{14\times14\times512}$
		\State Question feature: $f_{i} \in R^{1\times512}$
		\State $G_{c}:Conv2d(g_{i}), G_{c} \in R^{14\times14\times512}$
		\State  $F_{t}:Tile(f_{i}), F_{t} \in R^{14\times14\times512}$ 
		\State $h_{a}= \tanh(W_{I}G_{c} + W_{Q}F_{t}+b_{q})$
		\State $z_{a}= W_{a}h_{a} + b_{a}, z_{a}  \in R^{14\times14\times1} $
		\State $\alpha= \frac{\exp{z_i}}{\sum_{i=1}^{M}{\exp{z_i}}}, \alpha  \in R^{14\times14}$
		\State return $\alpha$
		\EndProcedure
	\end{algorithmic}
\end{algorithm}

 \begin{table*}[htb]
  \centering
  \begin{tabular}{|l|c|c|c|c|c|c|}
    \hline
    \textbf{Model}&  \textbf{R@1} & \textbf{R@5} & \textbf{R@10}& \textbf{MRR} & \textbf{Mean} & \textbf{NDGC}\\
    \hline
        SAN(Baseline)     & 40.93& 72.61& 83.50 &0.5616 & 5.11& 0.4595\\
        HCIAE     & 41.15& 72.67& 82.98 &0.5763 & 5.32& 0.4671\\
        MCB-att & 42.62& 74.45& 84.33 &0.6013 & 5.13& 0.4715 \\\hline
        GTA & 43.89 & 77.48& 87.21 &0.6045 & 4.81&0.4808 \\  
        GIA  &44.03 & 78.57  &88.60  &0.6092 & 4.68& 0.4877\\
        GMA\_cat  & 44.82& 80.05& 89.33 &0.6126 & 4.26 & 0.5012\\
        GMA\_MCB  & 45.10& 80.48& 90.15 &0.6187 & 3.91& 0.5095 \\ \hline
        GMA\_MCB-att   & \textbf{45.66}   & \textbf{81.62}  &\textbf{91.26}  &\textbf{0.6234} & \textbf{3.68} &  \textbf{0.5168}
        \\\hline
  \end{tabular}
  \caption{\label{tab_ablation} Ablation analysis of our  model on VisDial-v1.0 in test-std dataset.}
\end{table*}

\section{Experiments}{\label{experiments}}


We evaluate our proposed model in the following ways: first, we evaluate our proposed Granular Multi-modal Attention Network against other variants described in section~\ref{variants}. Second, we have shown rank correlation in table-~\ref{tab_rank_correlation} to analyze correlation between attention mask with the gradient class activation map. Third, we compare our method with state-of-the-art(SOTA) methods in table-~\ref{tab_SOTA} such as `visdial'~\cite{Das_CVPR2017}. Finally, we show the Grad-CAM ~\cite{selvaraju2017grad} visualization of aleatoric uncertainty and baseline model(late fusion). We further compare our model with state-of-the-art model such as `visdial'~\cite{Das_CVPR2017}. The quantitative evaluation is conducted using standard retrieval metrics namely models are evaluated on standard retrieval metrics – (1) mean rank, (2) recall @k and (3) mean reciprocal rank (MRR) of the human response in the returned sorted list.

\begin{table}[h]
  \centering
  \begin{tabular}{|l|c|c|c|}
    \hline
    \textbf{Model}&  \textbf{Rank Correlation}& \textbf{P value} & \textbf{EMD} \\
    \hline
        SAN&0.3415 &1.7913 &0.48 \\
        MCB &0.3326 &1.7973 &0.47 \\
        GTA& 0.3549 &1.7840 &0.44\\
        GIA &0.3590 &1.7814  & 0.42 \\
        GMA &\textbf{0.3670} &\textbf{1.7701} &0.41\\\hline
  \end{tabular}
  \caption{\label{tab_rank_correlation} Rank Correlation, EMD of Grad-CAM with attention mask}
\end{table}

    \subsection{Dataset}
    We evaluate our proposed approach by conducting experiments on  Visual Dialog dataset~\cite{Das_CVPR2017}, which contains human annotated questions based on images of MS-COCO dataset.  This dataset was developed by pairing two subjects on Amazon Mechanical Turk to chat about an image. One person was assigned the job of a ‘questioner’ and the other person acted as an ‘answerer’. The questioner sees only the text description of the an image which is present in caption from MS-COCO dataset and the original image remains hidden to the questioner. Their task is to ask questions about this hidden image to “imagine the scene better”. The answerer sees the image, caption and answers the questions asked by the questioner. The two of them can continue the conversation by asking and answering questions for 10 rounds at max. We have performed experiments on VisDial 1.0.
    Visual dialog v1.0 contains 123k dialogues on COCO-train for training split and 2k dialogues on Visual dialog val2018 images for validation split and 8k dialogues on visual dialog test-2018 for test-standard set.The caption is considered to be the first round in the dialog history. 



 \begin{table}[h]
  \centering
  \begin{tabular}{|l|c|c|c|}
    \hline
    \textbf{Model}&   \textbf{R@10}& \textbf{MRR} & \textbf{Mean} \\
    \hline
        GTA (K=8) & 86.50 &0.5616 & 5.41\\
       GTA (K=32) & 86.48 &0.5763 & 5.32\\
       GTA (K=64) & 87.21 &0.5813 & 4.81\\
       GTA (K=128) & 87.35 &0.5820 & 4.83\\\hline
        GIA (K=8) & 87.52 &0.5745 & 5.01 \\  
        GIA (K=32) &87.92  &0.5792 & 4.93\\
        GIA (K=64) & 88.60 &0.5926 & 4.68 \\
        GIA (K=128)& 88.63 &0.5943 & 4.23\\\hline
        GMA (K=8) & 89.02 &0.6026 & 4.26 \\
        GMA (K=32) & 89.15 &0.6187 & 3.91 \\
        GMA  (K=64)& 90.12 &0.6213 & 3.63\\
        GMA  (K=128)&\textbf{91.26}  &\textbf{0.6234} & \textbf{3.68} 
        \\\hline
  \end{tabular}
  \caption{\label{tab_ablation_granular} Ablation analysis of our  model with respect to various Granular features on VisDial-v1.0 in validation dataset.}
\end{table}

\begin{figure*}[!htb]
	\centering
	\includegraphics[width=1.0\textwidth]{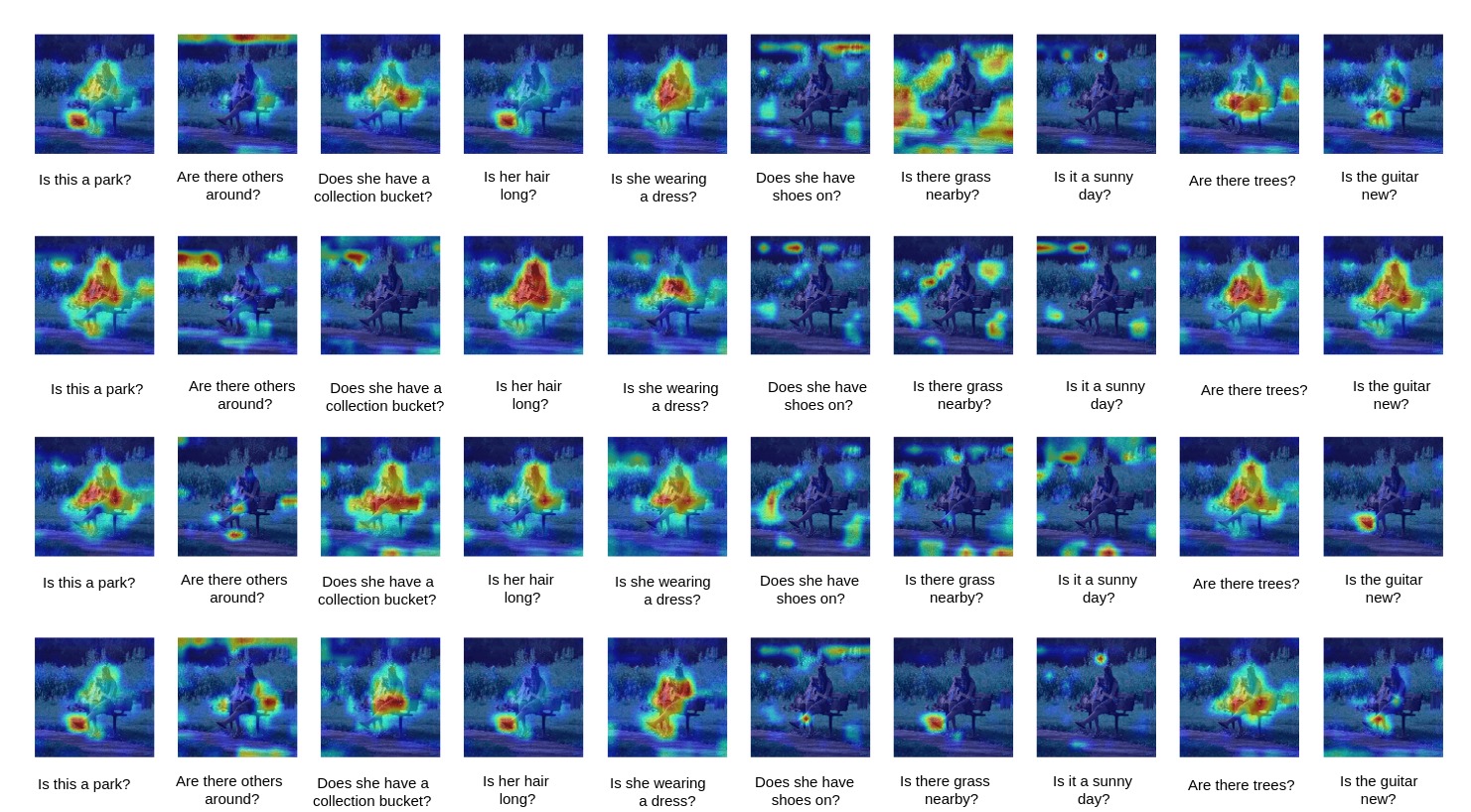}
	\caption{\textit{Visualization of attentions:} In this figure, the first row refers to SAN attention visualization of visual dialog model and second row refers to MCB-att attention visualization, third row refers to GIA attention visualization, fourth row refers to GTA attention visualization and fifth row refers to GMA attention visualization. The first column indicates the visualization of attention rounds of dialog from round 1 to 10.}
	\label{fig:Attention_Visualization}
	\vspace{-0.9em}
\end{figure*}

\subsection{Ablation Analysis on Granular Feature}\label{variants}
\begin{figure}[!htb]
	\centering
	\includegraphics[width=0.5\textwidth]{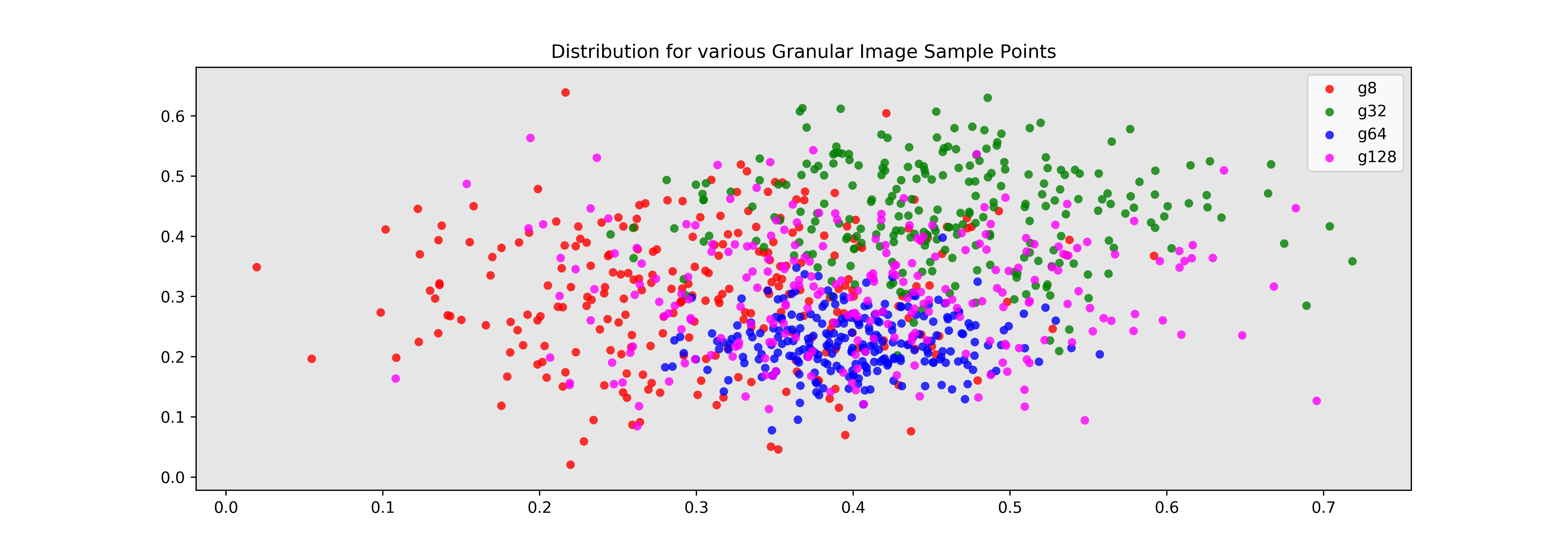}
	\caption{This figure shows the distribution of granular feature for various granular value such as K=8,32,64,128.}
		\label{fig:tsne}
\end{figure}

We conduct an experiment on various granular features for image and question attention map. We start with K=8,32,64,128 granular per image. We can observe that K=64, we observe significant improvement in accuracy score as shown in table-\ref{tab_ablation_granular}.  Further increasing K=128 it improves but not significant. So we select number of granular object is 64 in case of image and  also in question. Granular feature distribution for various value of K is shown in figure -\ref{fig:tsne}. From the distribution we can observe that feature are are correlated and lies between 0.2 to 0.6.  

\subsection{Ablation Analysis on Attention Network}\label{variants}
In this, we provide comparison of our proposed model GMA and other variants along with baseline model using various metrics in the table -~\ref{tab_ablation}. Each row provides results for one of the variations. The first block provides scores of our implementation for traditional methods such as stack attention method which is our baseline,  History-Conditioned Image Attentive Encoder (HCIAE)~\cite{lu2017best} method and Multimodal Compact Bilinear Pooling (MCB)~\cite{fukui2016multimodal} based attention method. The second block provides scores for our proposed Granular Image (GIA) and Text(GTA) attention model and the third block provides scores for different variant of our proposed multimodal attention model GMA. It is apparent that our best variant (GMA(MCB-att)) outperforms all the other variants achieving an improvement of 3.57\% in R@1 score, 4.22\% in R@5 score, 3.77\% in R@10 score, 0.0899\% in MRR and 2.27\% in mean over the baseline variant.

\begin{figure*}[!htb]
     \small
     \centering
     \begin{tabular}[b]{c c c c c}
        (a) K=0 & (b) K=8 & (c) K=32  & (d) K=64 & (e) K=128 \\ 
      \includegraphics[width=0.18\textwidth]{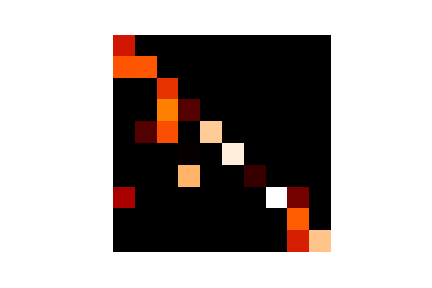}
     & \includegraphics[width=0.18\textwidth]{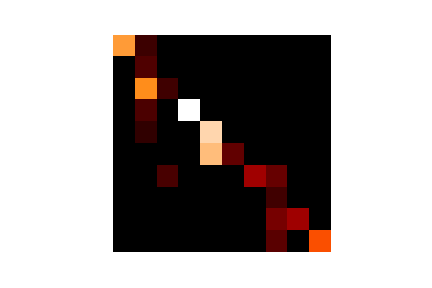}
     & \includegraphics[width=0.18\textwidth]{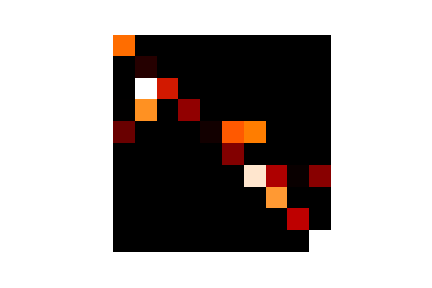}
     & \includegraphics[width=0.18\textwidth]{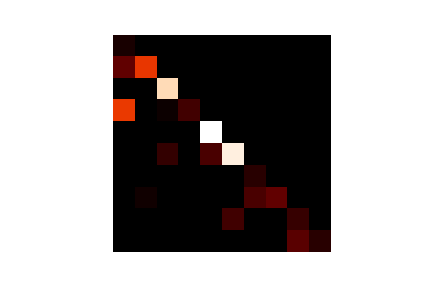}
      & \includegraphics[width=0.18\textwidth]{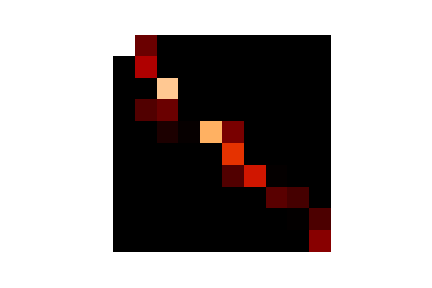}
       \end{tabular}
\caption{  \label{fig:ran_20} Joint probability distribution of GRAD-CAM mask and predicted attention mask distribution . We observe that as overlapping or EMD improves, the Joint Distribution is also improves as we increase K=8 to K=128. The Earth  Mover Distance between them is 0.37. We observe that as overlapping or EMD improves, the Joint Distribution is also improves.}
\vspace{-1.5em}
 \end{figure*}
 
 \begin{table*}[htb]
   \centering
   \begin{tabular}{|l|c|c|c|c|c|c|}
     \hline
     \textbf{Model}&  \textbf{R@1} & \textbf{R@5} & \textbf{R@10}& \textbf{MRR} & \textbf{Mean} & \textbf{NDGC}\\
     \hline
         Baseline & 40.56& 71.35& 82.83 &0.53 & 5.95 & 0.450\\
         HRE \cite{Das_CVPR2017}  &39.93&70.45 &81.50&0.54& 6.41 & 0.454 \\
         LF\cite{Das_CVPR2017}  &40.95& 72.45&82.83 &0.55 &5.95& 0.453\\
         MN \cite{Das_CVPR2017} &40.98& 72.30& 83.30 & 0.55&5.92& 0.475 \\
         MN-att \cite{Das_CVPR2017} &42.43& 74.00& 84.35 & 0.56&5.59& 0.476\\
         LF-att \cite{Das_CVPR2017}  &42.85& 74.83&85.05 &0.57 &5.41& 0.473 \\
         NMN \cite{kottur2018visual} & \textbf{47.50}& 78.12 &88.81& 0.61 &4.40 &\textbf{0.540} \\\hline
         GMA\_cat (ours) & 44.82& 80.05& 89.33 &0.61 & 4.26 & 0.5012\\
         GMA\_MCB (ours) & 45.10& 80.48& 90.15 &0.61 & 3.91& 0.5095 \\ 
         GMA\_MCB-att (ours)  &{45.66}   & \textbf{81.62}  &\textbf{91.26}  &\textbf{0.62} & \textbf{3.68} & {0.5168}\\ \hline
   \end{tabular}
   \caption{\label{tab_SOTA}SOTA results for Visual dialog v1.0 in Test-Standard .}
 \end{table*}
 
\subsubsection{Rank correlation, P value and EMD }\label{Correlation}

In the section, we measure Rank correlation(RC), P-value and Earth mover's distance (EMD). RC is a measure of monotonicity between two datasets and it is usually used for comparing images. P value indicates rough probability of uncorrelated system producing datasets having rank correlation equal to or more than those produced by these datasets.
We provide the quantitative ablation results for rank-correlation(higher is better) and P value(lower is better) in table -~\ref{tab_rank_correlation}. It is apparent that MCB outperforms other models in generating attention maps with rank correlation value of 0.3570 and p value of 1.7781 followed by GIA and GTA with rank correlation of 0.3509 and 0.3519 respectively and P value of 1.7820 and 1.7834. EMD is a measure of dissimilarity between two distributions. We also provide quantitative results for EMD in table -~\ref{tab_rank_correlation}. There is 7\% improvement over baseline(SAN) model.

\begin{figure}[ht]
	\centering
	\includegraphics[width=0.35\textwidth]{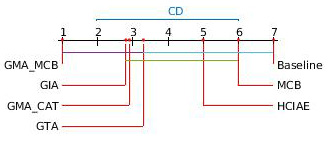}
	\caption{The mean rank of all the models on the basis of all scores are plotted on the x- axis. CD=4.0277, p=0.00008961. Here our GMA model and others variants are described in section~\ref{variants}. The colored lines between the two models represents that these models are not significantly different from each other.}
	\label{fig:SSA}
 	\vspace{-1.42em}
\end{figure}

\subsubsection{Statistical Significance Analysis}
We analyze Statistical Significance~\cite{Demvsar_JMLR2006} of our GMA model against other models mentioned in section~\ref{variants}. The Critical Difference (CD) for Nemenyi~\cite{Fivser_PLOS2016} test depends on given $\alpha$ (confidence level, which is 0.05 in our case) for average ranks and N(number of tested datasets). 
Low difference in ranks for two models implies that they are significantly less different. Otherwise, they are statistically different. Figure~\ref{fig:SSA} visualizes the post hoc analysis using the CD diagram. It is clear that GMA works best and is significantly different from other methods. Models within a single colored line are statistically indifferent.
\vspace{-0.7em}
\subsection{Comparison with Other Baselines}\label{baseline_sota}
\vspace{-0.7em}
The comparison of our method with various state-of-the-art methods for visual dialog dataset v1.0 are provided in table -~\ref{tab_SOTA}. The first block of the table consists of state-of-the-art methods for visual dialog model and second block consist of variant of our proposed method. Final row (GMA(MCB-att)) is our best proposed model. We compare our results with the baseline model `Late-fusion-QIH'~\cite{Das_CVPR2017}. We observe that we get an improvement of about 3.77\% \& 1.55\% in R@10 score and 9\% \& 5\% in MRR over the baseline model and the best model of Das et.al~\cite{Das_CVPR2017} respectively. We develop our GMA model on basic attention model (SAN \cite{Yang_CVPR2016}) and we achieve improvement of 5\% on MRR score. The simple and flexible  architecture of our GMA model can adapt on latest attention model also.  

 \vspace{-0.3em}
\subsection{Qualitative Result}\label{Results}
 \vspace{-0.7em}
We provide the qualitative results which can distinguish between baseline results with our GMA modal and other variants (SAN, GIA, GTA) as shown in Figure -~\ref{fig:Attention_Visualization} for a particular example. It is apparent that our proposed GMA model improves dialog attention probability over all other methods. For example, in the first image, the question was ``Is this in a park?". The attention mask for the  baseline model distributed over completer image, however our proposed model mainly focused on the field, plant and background image. Our proposed model explains about field, plant and background image, which provides extra information about the query, thus eventually we can observe that granularity in the image and question help to increase confidence of the attention map in the answer. The localisation of attention map increases from SAN to GMA as shown in Figure-\ref{fig:Attention_Visualization}. SAN is least localised and GMA is the most significant. Finally, GMA attention map is much more localised as compare to GTA, GIA model. 
\vspace{-0.85em}
\section{Conclusion}
\vspace{-0.7em}
In this paper we propose a novel Granular Multi-modal Attention Network that aims to jointly attend to appropriately sized granular image attention regions and granular textual regions using the correct context for each cue. We observe that the proposed attention regions provide improved attention regions as evaluated using a thorough empirical analysis. We further observe that the improved attention obtained using the proposed method consistently improves results for the task of visual dialog. Moreover, the proposed attention regions also correlate well with the regions as obtained by visualizing the gradients using Grad-cam. Thus, we consider that we are obtaining consistent attention regions that aid the network in solving the task of visual dialog. In future, we aim to further explore the proposed method in more such vision and language based tasks. We also aim to further explore the idea of obtaining correct semantic granular regions for solving various tasks.
{\small
\bibliographystyle{ieee_fullname}
\bibliography{egbib}
}

\end{document}